\documentclass[10pt,letterpaper]{article}
\usepackage[top=0.85in,left=2.75in,footskip=0.75in]{geometry}

\usepackage{amsmath,amssymb}

\usepackage{changepage}

\usepackage{textcomp,marvosym}

\usepackage{cite}

\usepackage{nameref,hyperref}

\usepackage[nopatch=eqnum]{microtype}
\DisableLigatures[f]{encoding = *, family = * }

\usepackage[table]{xcolor}

\usepackage{array}

\newcolumntype{+}{!{\vrule width 2pt}}

\newlength\savedwidth

\raggedright
\usepackage[aboveskip=1pt,labelfont=bf,labelsep=period,justification=raggedright,singlelinecheck=off]{caption}

\makeatletter
\renewcommand{\@biblabel}[1]{\quad#1.}
\makeatother

\usepackage{lastpage,fancyhdr,graphicx}
\usepackage{epstopdf}
\fancyheadoffset[L]{2.25in}
\fancyfootoffset[L]{2.25in}
\usepackage{graphicx}
\usepackage{subfigure}
\usepackage{array}
\usepackage{tabularx}
\usepackage{booktabs}
\usepackage{tikz}
\usepackage{amsmath}
\usepackage{float}

\begin{document}
\vspace*{0.2in}

\begin{flushleft}
{\Large
\textbf\newline{Supervised Learning without Backpropagation using Spike-Timing-Dependent Plasticity for Image Recognition} 
}
\newline
\\
Wei Xie\textsuperscript{1},
\\
\bigskip
\textbf{1} Department of Physics and Astronomy, Purdue University, 525 Northwestern Avenue, West Lafayette, IN 47907 USA
\bigskip

* wxie@purdue.edu

\end{flushleft}
\section*{Abstract}
This study introduces a novel supervised learning approach for spiking neural networks that does not rely on traditional backpropagation. Instead, it employs spike-timing-dependent plasticity (STDP) within a supervised framework for image recognition tasks. The effectiveness of this method is demonstrated using the MNIST dataset. The model achieves approximately 40\% learning accuracy with just 10 training stimuli, where each category is exposed to the model only once during training (one-shot learning). With larger training samples, the accuracy increases up to 87\%, maintaining negligible ambiguity. Notably, with only 10 hidden neurons, the model reaches 89\% accuracy with around 10\% ambiguity.  This proposed method offers a robust and efficient alternative to traditional backpropagation-based supervised learning techniques.


\section*{Introduction}
Image recognition has become ubiquitous, powering applications such as self-driving cars, medical image analysis, and facial recognition systems. Deep learning approaches \cite{deep_ANN} have achieved remarkable success, but their reliance on extensive computations requires powerful hardware, leading to high energy consumption and hindering deployment in devices with limited resources. Furthermore, deep learning models often lack biological plausibility, limiting their potential to understand neural computation and develop neuromorphic systems that mimic brain-like processing. Spiking neural networks (SNNs) \cite{SNN_ref} offer a bioinspired alternative to deep learning for image recognition. SNN neurons communicate through discrete spikes, mimicking the information processing of biological neurons. This spiking behavior promises lower power consumption and potentially better reflects the brain's processing capabilities. Spike Timing-Dependent Plasticity (STDP) \cite{stdp_ref} is a learning rule that allows SNNs to learn by dynamically adjusting synaptic strengths based on the timing of pre- and postsynaptic neuron spiking. This learning mechanism allows SNNs to extract features from the data, which makes them potentially well suited for image recognition tasks.

Various STDP learning rules and SNN architectures have been explored \cite{SNN_overview}.  Unsupervised learning, leveraging STDP, allows SNNs to autonomously acquire selectivity to recurring input patterns without external guidance. Supervised learning in SNNs using backpropagation is hindered by the nondifferentiable nature of spiking neurons and the "weight transport" problem \cite{SNN_w_transport_prob}.  While significant progress has been made in addressing these challenges, opportunities remain to develop more bioplausible solutions that better capture the complexities of biological systems. Beyond backpropagation, other supervised learning approaches include optimizing the probability of desired output spikes or implementing competitive dynamics among output neurons that represent distinct data classes. In terms of architectural design, researchers have proposed various SNN models, including deep-fully-connected SNNs, spiking convolutional neural networks, and spiking deep belief networks. These models have demonstrated performance comparable to traditional deep neural networks on a range of tasks. 

This paper presents a supervised learning approach for SNNs that integrates gradient-free optimization with STDP for image classification on the MNIST dataset ~\cite{MNIST}. To enhance class selectivity, each hidden-layer neuron is assigned to a unique group that corresponds to a specific digit class. Synaptic weights are updated only when the target class aligns with the neuron's group, promoting specialization. A unique feature of the proposed approach is the absence of inhibitory synaptic connections between neurons during the training stage.   This allows neurons to fully explore the input space,  fostering the development of class preferences. Lateral inhibition, introduced only during validation and testing, enables competitive dynamics among neurons, leading to a clean classification based on firing rates. The network architecture is scalable, supporting the construction of deep SNN models for complex tasks. The experimental results on MNIST demonstrate the efficacy of the proposed method in achieving good classification accuracy. The SNN simulations are performed using the Brian2 simulator v2.7.1 \cite{brian2} \footnote{Available on GitHub: "https://github.com/wxie2013/GDFree-supervised-SNN-MNIST"} on HPC resources provided by the Purdue Rosen Center for Advanced Computing \cite{McCartney2014}. 

The remaining sections of this paper are organized as follows. Section 2 presents the neuron and synapse models, as well as the network architecture. Section 3 details the implementation of supervision in the STDP learning rule. Section 4 describes the hyperparameter tuning process. Section 5 evaluates the performance of the network on the MNIST dataset. Section 6 concludes the paper.

\section*{Network Design}
\label{network}
\subsection*{Modeling Neurons and Synapses}
The leaky integrate-and-fire model is chosen as a computational framework to simulate neuronal dynamics. Neurons are simplified as electrical circuits, incorporating capacitance and resistance, while essential features are captured without undue complexity. The membrane potential (V) evolves over time according to the following differential equation:
\begin{equation}
\tau_{m}\frac{dV}{dt} = (E_{\text{rest}} - V) + g_{\text{e}}(E_{\text{exc}} - V) + g_{\text{i}}(E_{\text{inh}} - V)
\label{eq_neuron}
\end{equation}
where $E_\text{rest}$ represents the resting membrane potential, $E_\text{exc}$ and $E_\text{inh}$ denote the equilibrium potentials of the excitatory and inhibitory synapses, respectively, $g_\text{e}$ and $g_\text{i}$  correspond to the conductances of the excitatory and inhibitory synapses, and $\tau_{m}$ signifies the time constant of neurons.
When a neuron's membrane potential exceeds its threshold, it generates an action potential and subsequently resets its potential to $V_\text{reset}$. The threshold, initially set at $V_\text{thres}$, is dynamically adjusted based on the neuron's firing rate. During the subsequent refractory period, neurons can not spike again.  Synaptic strengths are modulated by changes in conductance. When a presynaptic spike occurs at the synapse, synapses increase their conductance by synaptic weight $\textit{w}$. However, when the neuron is not active, the conductance decays exponentially. This dynamic can be mathematically described as
\begin{equation}
\tau_{g_e}\frac{dg_e}{dt} = -g_e, \quad  
\tau_{g_i}\frac{dg_i}{dt} = -g_i 
\label{eq_syn}
\end{equation}
where $\tau_{g_e}$ and $\tau_{g_i}$ represent the time constant of the excitatory and inhibitory conductance of a  postsynaptic neuron, respectively.  

Table \ref{tab:neuron_syn_pars} presents the fixed parameters of the neuron and synapse model. These parameters remain constant for neurons and synapses in different layers of the network. Other parameters, such as $\tau_m$, $\tau_{g_e}$, $\tau_{g_i}$ are determined through an optimization process as described in Sect. \ref{hyperpar_tune}. 
\begin{table}[ht]
    \centering
    \begin{tabular}{c|c|c|c|c}
        \hline
        $E_{exc}$  & $E_{inh}$  &  $E_{rest}$  & $V_{thres}$  & $V_{rest}$ \\
        \hline
        0 ($\it{m}$V)& -100 ($\it{m}$V) & -65 ($\it{m}$V) & -52 ($\it{m}$V)& -65 ($\it{m}$V)\\
        \hline
    \end{tabular}
    \caption{Parameters of hidden-layer neuron and synapse models that remains constant across all layers of a network.}
    \label{tab:neuron_syn_pars}
\end{table}

\subsection*{Adaptive Threshold}
Balanced neuronal activation is essential for optimal network performance\textbf{.} To prevent individual neurons from unduly dominating network responses, an adaptive membrane threshold mechanism, inspired by the work in \cite{adpt_thres1,adpt_thres2},  is implemented.  The adaptive threshold ($V_t$) is initialized at $V_{thres}$ and increases by a small value upon each neuronal firing. The small incremental change is calculated as $\Delta V_t \Theta_{vt}$, where $\Theta_{vt}$ prevents the threshold of a neuron from reaching too high and stops firing. 
\begin{equation}
    \Theta_{vt} =  \left( 0.5 - 0.5 \tanh \left(\left( \frac{-2(V_t - V_{thres}(V_{tshift} - 0.5))}{V_{thres}} + 1 \right) / V_{tscale} \right) \right)
    \label{vt_sat}
\end{equation}
Here $\Delta V_t$ is the maximum potential increase upon each neuronal firing,  $V_{tshift}$ represents the point at which  $\Theta_{vt}$ start to decrease significantly from one,  and  $V_{tscale}$ regulates the  rate of decrease.  This formulation prevents excessive threshold elevation, thus maintaining neuronal responsiveness. When a neuron is not active, $V_t$ decreases exponentially as a function of time with a time constant $\tau_{adpt}$,  as described by: 
\begin{equation}
    \tau_{adpt}\frac{dV_t}{dt} = (V_{thres}-V_t)
    \label{eq_adpt}
\end{equation}

\subsection*{Network Architecture}
\begin{figure}[!ht]
    \centering
    \subfigure[]{
        \includegraphics[width=0.3\linewidth]{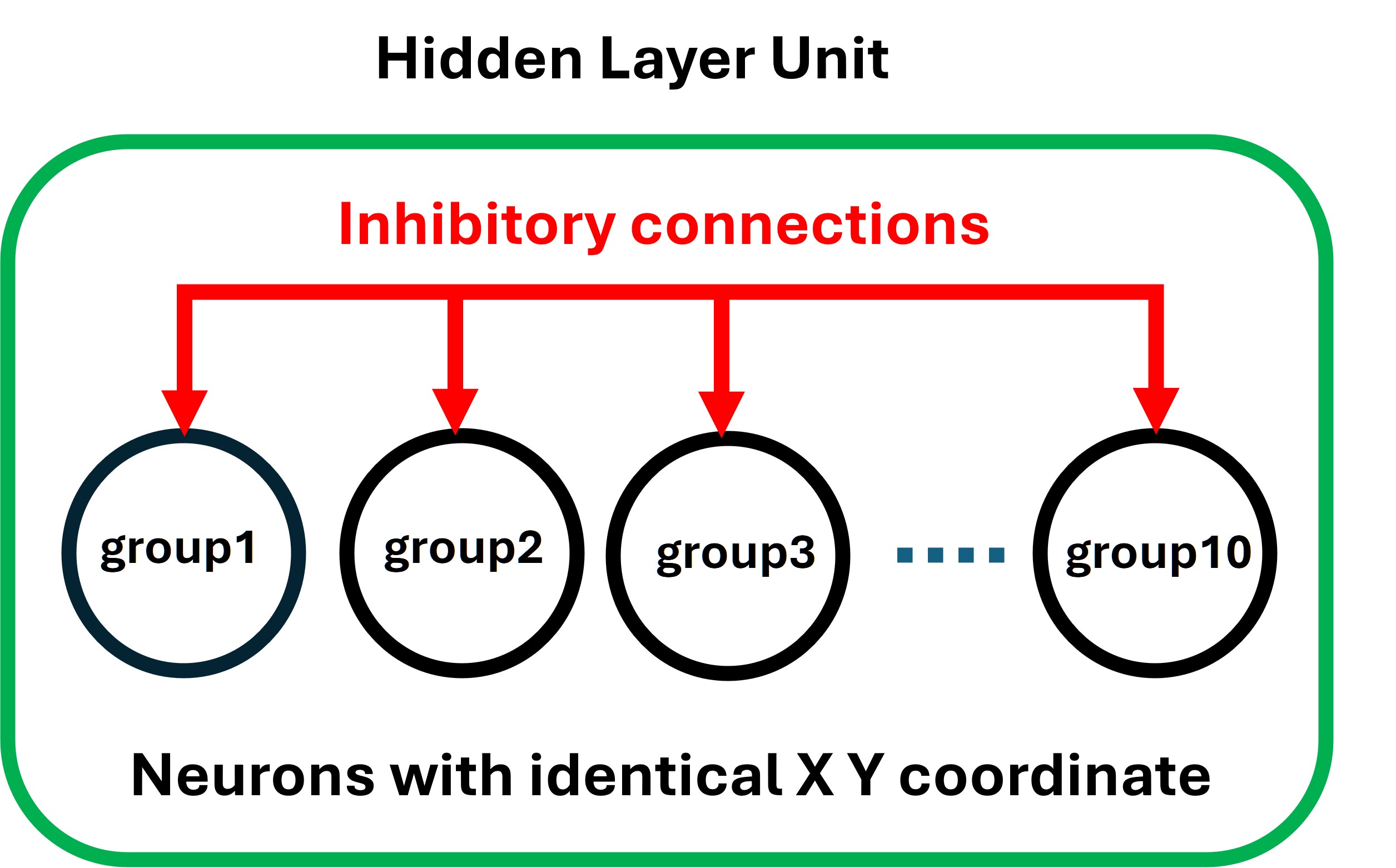}
    }
    \subfigure[]{
        \includegraphics[width=0.7\linewidth]{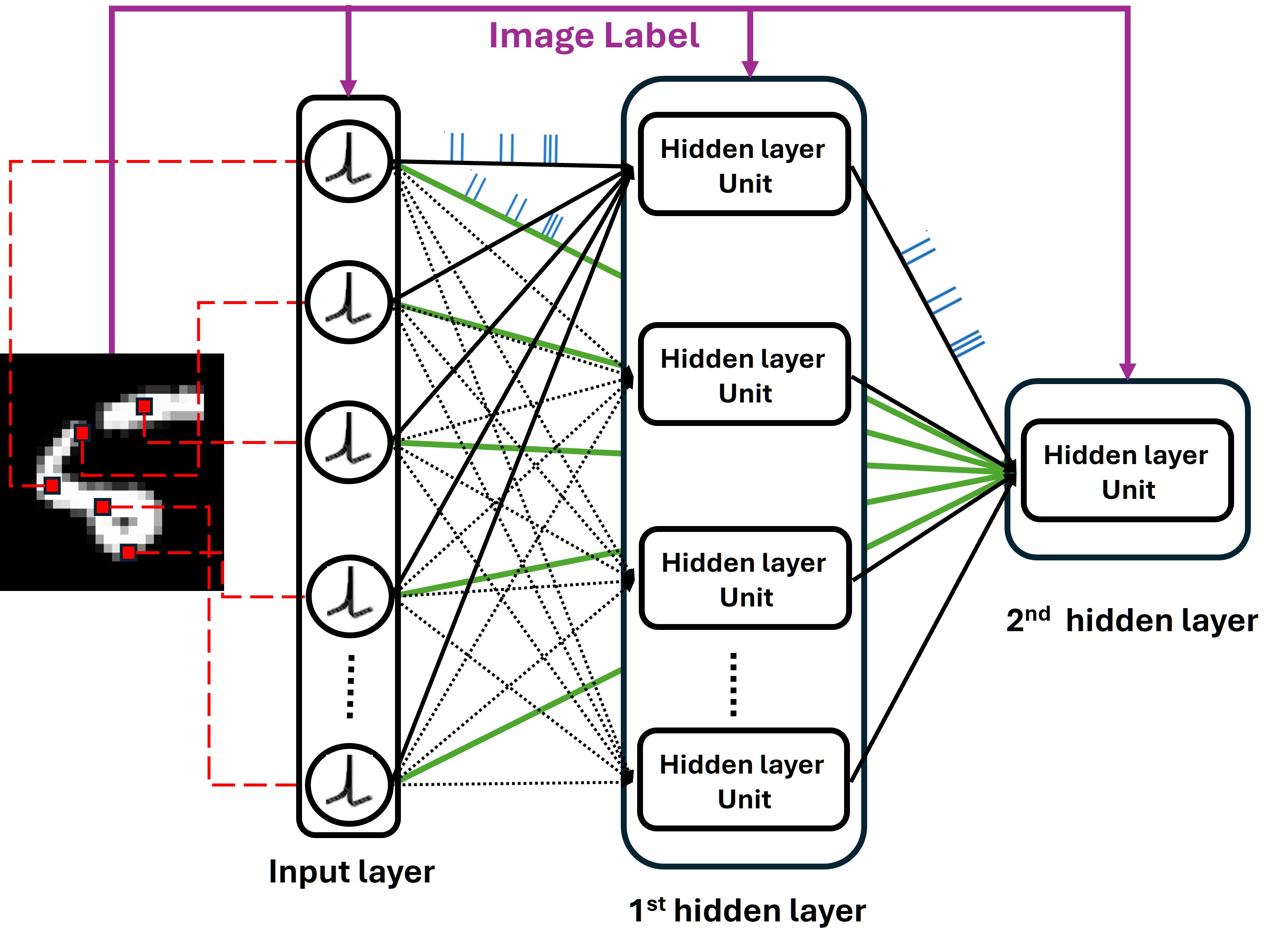}
    }
    \caption{(a) A single hidden layer unit comprising ten neurons, each dedicated to a specific digit between 0 and 9.  Neurons in a unit share identical spatial coordinates and are fully connected with inhibitory synapses during validation and testing. (b) An example of a network with two hidden layers. The input layer encodes image information through Poisson-distributed spike trains, with firing rates proportional to pixel intensities. Full connectivity exists between the input layer and the first hidden layer, as well as between the first and second hidden layers. During training, the image label is provided as supervisory information to each excitatory neuron.}
    \label{fig:network_arch}
\end{figure}
The network architecture is a hierarchical feedforward structure composed of an input layer and hidden layers, as illustrated in Figure \ref{fig:network_arch}. The input layer comprises 784 neurons, each representing a pixel of a MNIST image. These neurons encode image information through Poisson-distributed spike trains, with firing rates directly proportional to pixel intensities. Each hidden layer is divided into ten groups of neurons, with each group dedicated to recognizing a specific class of digit. A hidden layer unit, as depicted in Figure \ref{fig:network_arch}(a), consists of ten neurons, one for each digit group. All neurons within a unit share identical spatial coordinates (X, Y) positioned uniformly above the input layer. For this study, a fully connected architecture is employed, where the first hidden layer is fully connected to the input layer, and subsequent hidden layers maintain full connectivity to both the input layer and all preceding layers. The spatial coordinates assigned to neurons offer potential for future enhancements, such as localized connectivity between layers. 
To enhance digit selectivity and inter-digit competition during testing and validation, inhibitory connections are established between neurons belonging to different digit groups. The firing of a neuron in one group suppresses the activity of neurons in other groups, promoting focused digit representation. To facilitate the development of digit preferences by all neurons during training, these inhibitory connections are absent in the training phase.
The image classification is determined by the group exhibiting the highest firing rate in the last hidden layer.

\section*{STDP with Supervised Learning}
Excitatory synaptic weights are adjusted using the triplet STDP learning rule \cite {triplet_STDP}, known for its superior performance as demonstrated in \cite{Diehl_model}. According to this learning rule, a presynaptic spike at time $t^{pre}$ introduces a weight change:
\begin{equation}
     w(t) \rightarrow w(t) - o_1(t)[A^-_2 + A^-_3 r_2(t - \varepsilon)] \hspace{0.2cm} \textrm{ if } \hspace{0.2cm} t = t^{pre} 
     \label{stdp_pre}
\end{equation}
Conversely, a postsynaptic spike at time $t^{post}$ results in a weight change: 
\begin{equation}
     w(t) \rightarrow w(t) + r_1(t)[A^+_2 + A^+_3 o_2(t - \varepsilon)] \hspace{0.2cm} \textrm{ if } \hspace{0.2cm} t = t^{post}
     \label{stdp_post}
\end{equation}
Here,  $r_1$ and $r_2$ are presynaptic event detectors, while  $o_1$ and $o_2$ are postsynaptic detectors, each with biological counterparts.  $A^+_2$ and $A^-_2$ quantify the weight changes for pre-post pair or a post-pre pair, respectively. $A^+_3$ and $A^-_3$ represent the amplitude of the triplet term for potentiation and depression. For simplicity, the minimal model ($A^+_2$ = 0, $A^-_3$ = 0), which is intended to fit the visual cortex data, is used for this study.  

The supervison is incorporated  by modifying eq.(\ref{stdp_post}) as follows: 
\begin{equation}
    \centering
      w(t) \rightarrow w(t) + r_1(t) A^+_3 o_2(t - \varepsilon) \Theta_w (1 - \beta)
\end{equation}
where $\beta$ = 0 if the postsynaptic neuron group index matches the stimulus label, otherwise $\beta$ = 1. The variable $\Theta_w$ prevents premature weight saturation, and is defined as: 
\begin{equation}
    \Theta_w = \left( 0.5 - 0.5 \tanh \left(\left( \frac{2(w + w_{max}(w_{shift} - 0.5))}{w_{max}} - 1 \right) / w_{scale} \right) \right)
    \label{w_sat}
\end{equation}
with $w_{shift}$ determining the oneset of a significant decrease in $\Theta_w$ start from one,  and  $w_{scale}$ regulates the  rate of decrease.  

To promote equitable participation of neurons in network dynamics \cite{weight_norm},  the average weights of a postsynaptic neuron are normalized to $\lambda w_{max}$  with $w_{max}$ being the maximum possible weight.  This normalization is achieved by scaling each synaptic weight according to the following formula: 
\begin{equation}
   w \rightarrow w \cdot \frac{\lambda w_{\max} }{w_{\text{tot}} / N_{\text{pre}}}
   \label{w_scale}
\end{equation}
where $w_{tot}$ is the sum of all synaptic weights of a postsynaptic neuron and $N_{pre}$ is the number of presynaptic neurons connected to it. The hyperparameter $\lambda$ ($0\leq\lambda\leq1$) controls the degree of normalization of the weight. This procedure allows synaptic weight to have ample space to grow with a large sample of stimuli and is performed after processing each stimulus.  

\section*{Hyperparameter Optimization}
\label{hyperpar_tune}

\begin{table}[h!]
    \begin{tabularx}{\textwidth}{>{\centering}c|>{\centering}m{2.1cm}|>{\centering}m{2.1cm}|>{\small\centering\arraybackslash}X}

        \hline
        Par name  &  search range (base)  &  search range (2-hidden-layer)   & Description\\
        \hline
        $\tau_{adpt} (\textrm{{ms}}) $ & $ 10.0 - 10^8$       & $\textsc{n/a} $ &  Time constant of adaptive threshold eq.($\ref{eq_adpt}$). \\
        $\Delta V_t (\textrm{mV})    $ & $ 10^{-3} - 10^{-1}$ & $\textsc{n/a} $ &  Maximum increment of neuron threshold in eq.($\ref{vt_sat}$). \\
        $\tau_m (\textrm{ms})        $ & $ 10.0 - 200.0$      & $\textsc{n/a} $ &  Time constant of neuron membrane potential.\\
        $\tau_{g_e} (\textrm{ms})    $ & $ 1.0 - 10.0$        & $\textsc{n/a} $ &  Time constant of the excitatory conductance.\\
        $\tau_{g_i} (\textrm{ms})    $ & $  1.0 - 10.0$       & $\textsc{n/a} $ &  Time constant of the inhibitory conductance.\\
        $w_{i \rightarrow e} (max)   $ & $ 0.1 - 100.0$       & $\textsc{n/a} $ &  Maximum weight of a synapse between the input and a hidden layer.\\
        $\lambda_{i \rightarrow e}   $ & $ 10^{-3} - 0.5$     & $\textsc{n/a} $ &  Scaling factor in eq.($\ref{w_scale}$) for a synaptic weight between the input and a hidden layer \\
        $w_{e \rightarrow e} (max)   $ & $ \textsc{n/a}$      & $1.0 - 100.0  $ &  Maximum weight of a synapse between two hidden layers.\\
        $\lambda_{e \rightarrow e}   $ & $ \textsc{n/a}$      & $0.1 - 0.5    $ &  Scaling factor in eq.($\ref{w_scale}$) for a synaptic weight between two hidden layers\\
        $\Delta w_{e \leftrightarrow e}$& $10^{-2} - 100.0$   & $\textsc{n/a} $ &  Weight of the lateral inhibitory synapse among neurons in the same hidden layer. \\
        $max_{delay} (\textrm{ms})   $ & $0.0 - 200.0    $    & $0.0 - 200.0  $ &  Maximum delay of excitatory synapses. \\
        $A_2^-                       $ & $10^{-5} - 10^{-2}$  & $\textsc{n/a} $ &  $A_2^-$ in STDP learning rule eq.($\ref{stdp_pre}$) \\
        $V_{tscale}                  $ & $10^{-3} - 1.0$      & $\textsc{n/a} $ &  $V_{tscale}$ in eq.($\ref{vt_sat}$) \\
        $V_{tshift}                  $ & $0.0 - 1.0   $       & $\textsc{n/a} $ &  $V_{tshift}$ in eq.($\ref{vt_sat}$) \\
        $w_{scale}                   $ & $10^{-3} - 1.0 $     & $\textsc{n/a} $ &  $w_{scale}$ in eq.($\ref{w_sat}$)\\
        $w_{shift}                   $ & $0.0 - 1.0   $       & $\textsc{n/a} $ &  $w_{shift}$ in eq.($\ref{w_sat}$)\\
        \hline
    \end{tabularx}
    \newline
    \caption{Tuning ranges of hyperparameters with different network configurations. Base represent a network with one hidden layer which consists of one hidden layer unit. 2-hidden-layer represents a network with two hidden layers.}
    \label{tab:hyperpar_tune}
\end{table}

Hyperparameter optimization was conducted using the Hyperopt algorithm \cite{hyperopt} within the Ray framework \cite{liaw2018tune}. Each stimulus was presented to the network for 0.5 seconds. The input layer converted images into Poisson spike trains, starting with a mean firing rate proportional to 25\% of the original MNIST image intensity. To ensure sufficient spiking activity, the firing rate was adaptively increased by increments of 25\% of the intensity until the spike count in each hidden layer reached a minimum of five, or the full intensity was attained. Preliminary experiments indicates that image recognition accuracy  is relatively insensitive to this intensity modulation. 
Table \ref{tab:hyperpar_tune} outlines the hyperparameter search ranges for different network configurations. A base network consists of a single hidden layer with one hidden layer unit, while a 2-hidden-layer network comprises two hidden layers. 

\begin{table}[h!]
    \centering
    \begin{tabular}{c|c|c}
        \hline
        Par name  &  base   &  2-hidden-layer ($1^{st}/2^{nd}$) \\
        \hline
        $\tau_{adpt} (\textrm{{ms}}) $ & $10^6$       & $10^6$/$10^6$  \\
        $\Delta V_t (\textrm{mV})    $ & $4.4\times10^{-3}$     & $4.0\times 10^{-4}/3.0\times 10^{-3}$  \\
        $\tau_m (\textrm{ms})        $ & $200.0     $ & 170.0/190.0  \\
        $\tau_{g_e} (\textrm{ms})    $ & $0.4       $ & 1.0/0.3  \\
        $\tau_{g_i} (\textrm{ms})    $ & $4.0       $ & 3.0/3.0  \\
        $w_{i \rightarrow e} (max)   $ & $29.0        $ & 58.0/72.0  \\
        $\lambda_{i \rightarrow e}   $ & $0.28      $ & 0.34/0.24  \\
        $w_{e \rightarrow e} (max)   $ & $\textsc{n/a}$ & 100.0 \\
        $\lambda_{e \rightarrow e}   $ & $\textsc{n/a}$ & 0.15   \\
        $max_{delay}^{e \rightarrow e} (\textrm{ms})   $ &  $\textsc{n/a}$   & 50.0  \\
        $max_{delay}^{i \rightarrow e} (\textrm{ms})   $ &  0.0      & 10.0/0.0  \\
        $\Delta w_{e \leftrightarrow e}$ & 0.64   &  1.4/2.1\\
        $V_{tscale}                  $ &  0.18     & 0.21/0.21  \\
        $V_{tshift}                  $ &  0.10        & 0.40/0.40  \\
        $w_{scale}                   $ &  0.23     & 0.14/0.14  \\
        $w_{shift}                   $ &  0.30     & 0.40/0.40  \\
        \hline
    \end{tabular}
    \newline
    \caption{A set of hyperparameters demonstrating good validation efficiency, selected from multiple sets with comparable performance, for both base and a 2-hidden-layer networks.}
    \label{tab:best_par}
\end{table}

Hyperparameter optimization  was performed on a subset of 10,000 MNIST images for a single epoch and validated with a subset of 1000 MNIST images.  For the optimization of a multilayer network, all  hyperparameters except for  $\lambda_{e \rightarrow e}$, $w_{e \rightarrow e} (max)$ and $max_{delay}$,  were randomly selected from  sets  validated to have good performance during base network optimization. The parameters $\lambda_{e \rightarrow e}$ and $w_{e \rightarrow e} (max)$ can be influenced by the size of the preceding hidden layers, while the $max_{delay}$ between hidden layers may depend on the $max_{delay}$ between hidden layer and input layer.  In this study, the sizes of all hidden layers in a multilayer network are fixed at 10 neurons and the optimized  $\lambda_{e \rightarrow e}$, $w_{e \rightarrow e} (max)$ and $max_{delay}$ are thus tailored for this structure.   Multiple hyperparameter sets yielded comparable validation performance under these conditions and Table \ref{tab:best_par} presents one set of optimal hyperparameters identified in this study.  It is acknowledged that  hyperparameter with better performance may be attainable with access to greater computational resources.  

\section*{Results and Discussion}
The performance of two different network training methods was evaluated using $10^4$ testing samples. The direct approach involves training the network on a fixed number of training samples and then testing it as is. The disadvantage of this method is that the training time is proportional to the number of excitatory neurons in the hidden layers. As the number of neurons increases, the training process can become very slow. To overcome this limitation, we leverage the fact that the smallest network in this model, referred to as the base network, consists of only 10 neurons, each responsible for identifying a single MNIST label, with no lateral excitatory connections between them. During training, a larger network with, for example, one hidden layer containing 250 hidden neurons, can be decomposed into 25 base networks, each processing a different subset of training samples via a single computing process (worker). During testing, the 25 base networks are combined, applying inhibitory connections within each base network among neurons belonging to different groups. Each base network reads in its own respective training synaptic weights and neuron threshold. This approach significantly reduces training time, as each worker process only trains a smaller network. Furthermore, each small network uses a unique set of hyperparameters that have been validated to have good performance during the optimization stage, resulting in a combined network with a diverse set of hyperparameters.  This approach is termed  "parallel training with diversity". 
\begin{figure}[!ht]
    \centering
    \includegraphics[width=0.7\linewidth]{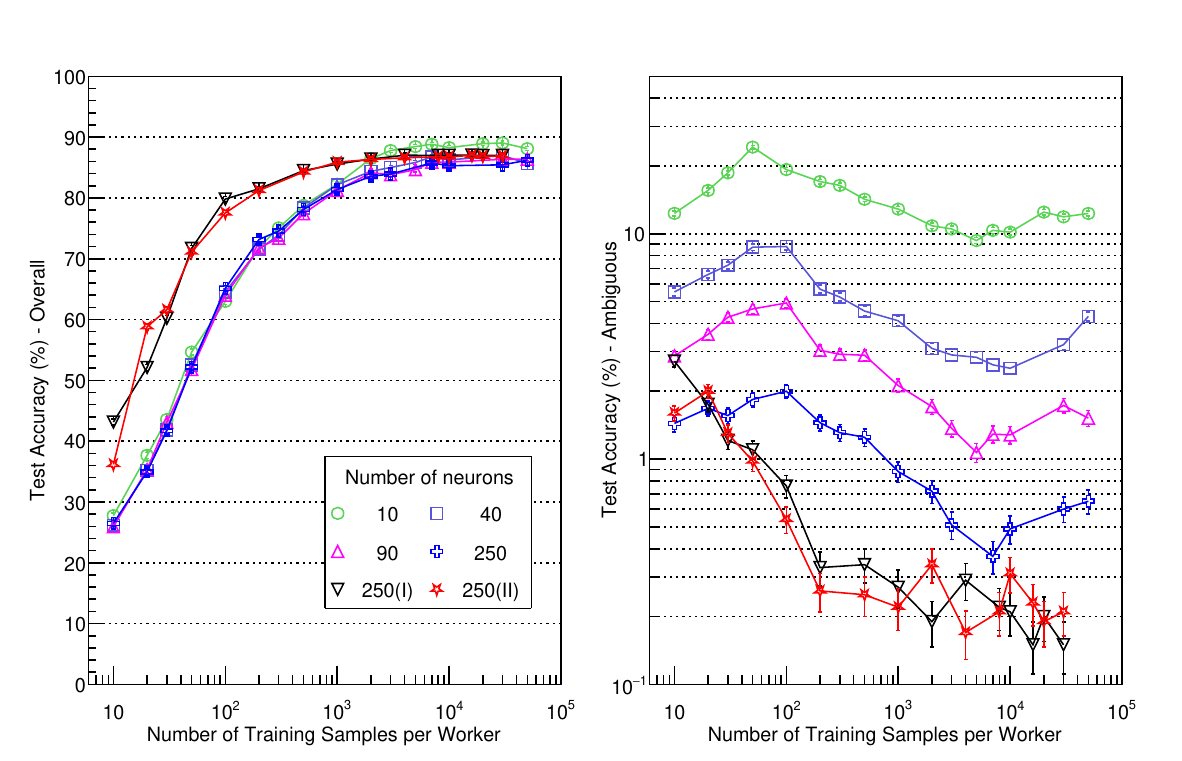}
    \caption{The influence of training sample size per worker on (left) overall test accuracy and (right) test accuracy with ambiguity for  networks with different hidden layer sizes. Each color represents a specific number of excitatory neurons in the hidden layer. The size of test sample is $10^4$ and the bars associated with  data points represent  statistical uncertainties. }
    \label{fig:eff_vs_nsample}
\end{figure}

Figure~\ref{fig:eff_vs_nsample} presents testing accuracy results for networks trained using parallel training with diversity and direct training approaches.   The data points of different colors represent the results with different network size and training approaches.  The  spike counts from all neurons within each neuron group in the last hidden layer are summed, and the group with the maximum spike counts is treated as the prediction of the network.  For some stimuli, more than one neuron groups have the same maximum spike counts.  In this case, if one of these groups matches the input label,  it is counted as a correct identification, albeit   with ambiguity.  
The left panel of the figure illustrates  the dependence of the overall testing accuracy on size of the training sample per worker,  including those identified with ambiguity.  The right panel displays the fraction of correctly identified stimuli with ambiguity, i.e., test accuracy with ambiguity.  The result labeled “250 (I)” comes from a network with one hidden layer containing 250 excitatory neurons, trained using parallel training with diversity. The result labeled “250 (II)” is from a network with two hidden layers, each containing 250 excitatory neurons, also trained using parallel training with diversity.  In both cases, 25 workers are used during the training stage. The other results are from networks of various sizes trained using the direct training approach.  Using the direct approach, the overall accuracy goes from $\sim 26\%$ to $\sim 89\%$ as the size of the training sample grows from 10 to 5$\times 10^4$.  Increasing the number of excitatory neurons in the hidden layer from 10 to 250 does not improve the overall test accuracy but  reduce  ambiguity by more than 10 folds. For smaller training samples, parallel training with diversity significantly outperforms the direct approach. As the training set size increases, their accuracies converge. However, the test accuracy with ambiguity using parallel training with diversity is significantly lower with the same network configuration. 
Compared to the single hidden layer network (250(I)), adding a second hidden layer with the specified configuration (250(II)) does not enhance the performance. This suggests that the information processed from the first hidden layer is too weak to influence the decision-making of the second layer. Other configurations, such as increasing the number of neurons in the first hidden layer,  introducing lateral connections,  or localized feedforward synaptic connections, may improve the network performance, as demonstrated for artificial neuron networks.  

\begin{figure}[!ht]
    \centering
    \includegraphics[width=0.7\linewidth]{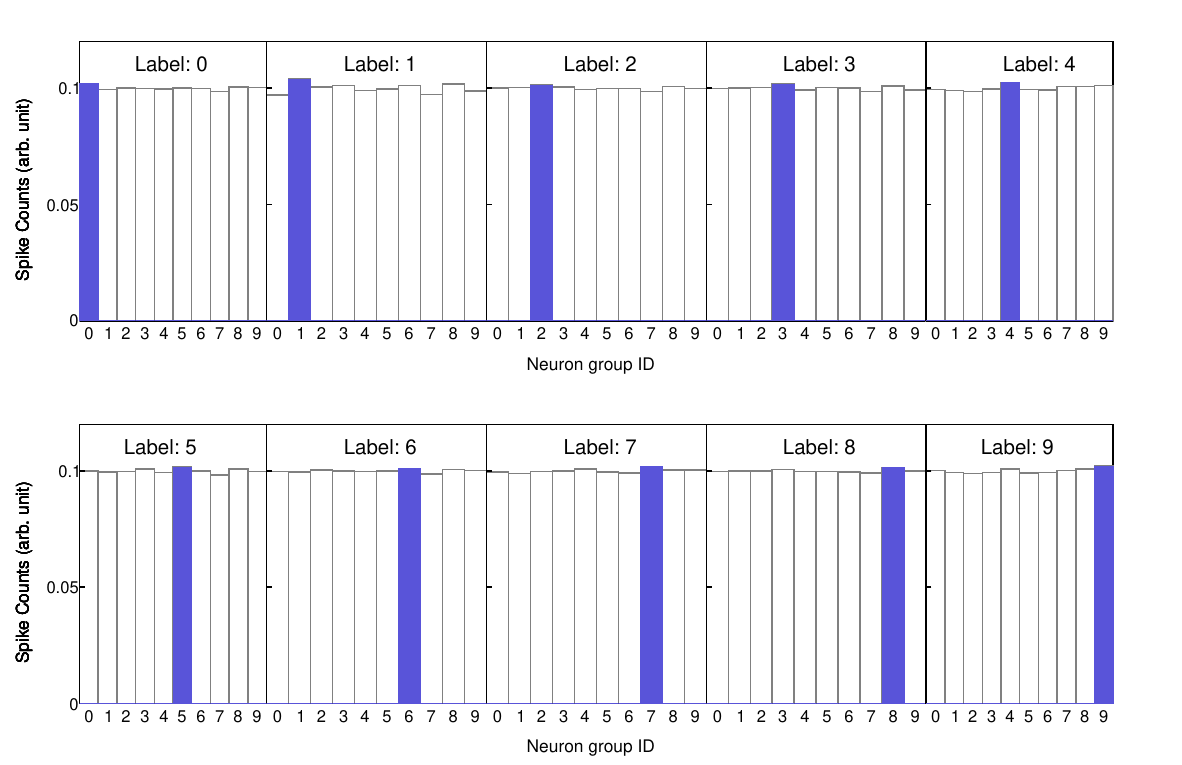}
    \caption{Spike count distribution of each neuron group for correctly identified test samples  across different input labels (highlighted in blue). The network has a single hidden layer with 250 neurons (equivalent to 25 workers) employing parallel training with diversity. Each worker processed 10 training stimuli. The size of test sample is $10^4$.}
    \label{fig:pre_10}
\end{figure}
\begin{figure}[!ht]
    \centering
    \includegraphics[width=0.7\linewidth]{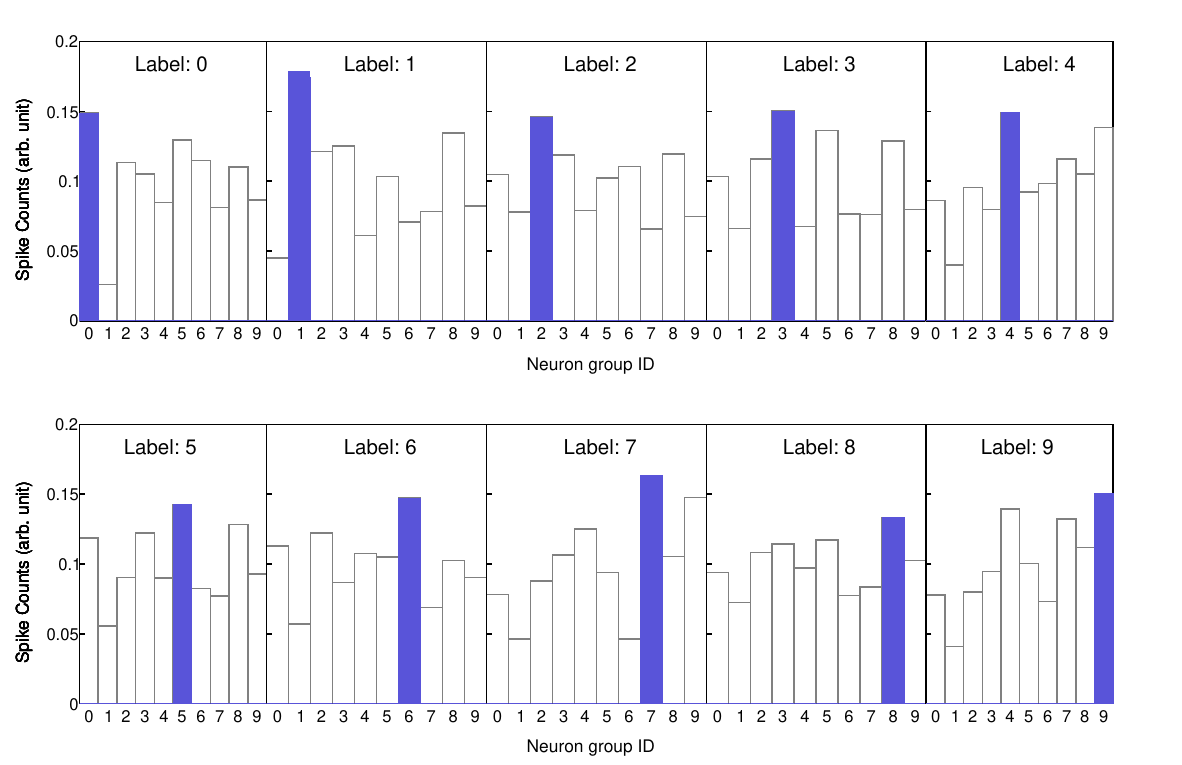}
    \caption{Spike count distribution of each neuron group for correctly identified test samples  across different input labels (highlighted in blue). The network has a single hidden layer with 250 neurons (equivalent to 25 workers) employing parallel training with diversity. Each worker processed $10^4$ training samples. The size of test sample is $10^4$.}
    \label{fig:pre_1e4}
\end{figure}

Figure~\ref{fig:pre_10} illustrates the spike counts from each neuron group in response to a specific label in the test sample, obtained through parallel training with diversity when the number of training samples per worker is 10. The network comprises of 250 neurons in the single hidden layer, with 25 workers (denoted as 250(I) in Figure~\ref{fig:eff_vs_nsample}). In this setup, each training stimulus is exposed to the network only once on average.  Due to the limited number of training samples, the synaptic weights have not fully specialized, leading to similar responses to a given stimulus across all neuron groups. Notably, even with such small differences, the network achieves a $\sim 40\%$ testing accuracy with low ambiguity. In contrast, Figure~\ref{fig:pre_1e4} displays the spike count distribution with a significantly larger training sample size per worker ($3\times10^4$). With this extensive training set, the synaptic weights of different neuron groups have become more specialized. Consequently, the spike rate of the correct neuron group is substantially higher than that of other groups, indicating improved discrimination. 

\begin{figure}[!ht]
    \centering
    \includegraphics[width=0.7\linewidth]{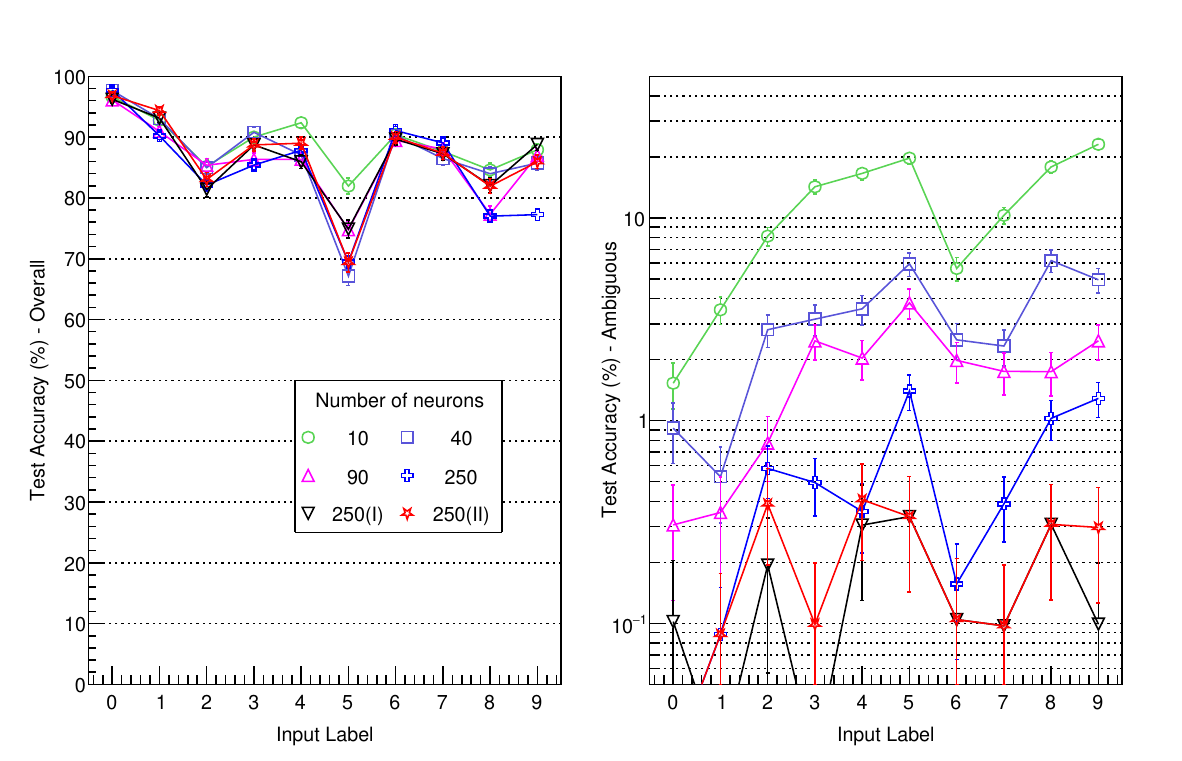}
    \caption{The influence of input labels on (left) overall test accuracy and (right) test accuracy with ambiguity for networks with different hidden layer sizes. Each color represents a specific number of excitatory neurons in the hidden layer. The size of test sample is $3\times10^4$ and the bars associated with  data points represent  statistical uncertainties.}
    \label{fig:eff_vs_label}
\end{figure}
\begin{figure}[!ht]
    \centering
    \includegraphics[width=0.7\linewidth]{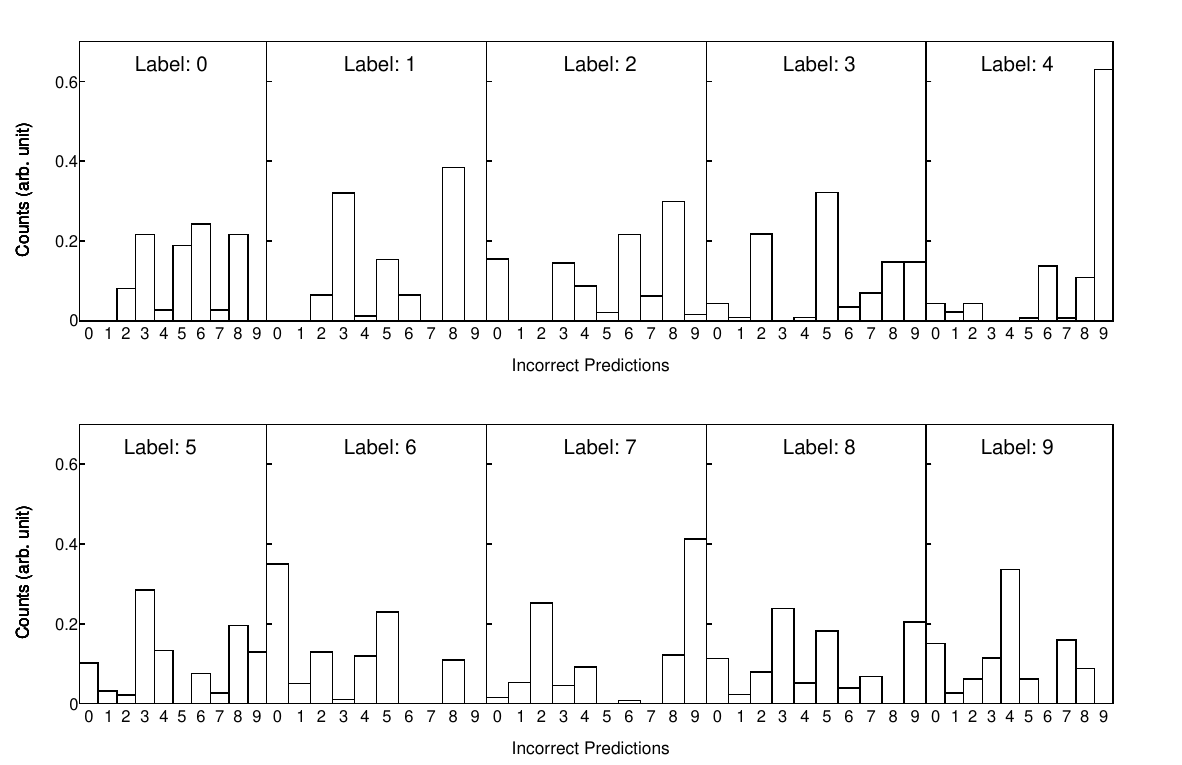}
    \caption{Distribution of incorrect predictions across different input labels. The network has a single hidden layer with 250 neurons (equivalent to 25 workers) employing parallel training with diversity. Each worker processed $3\times10^4$ training samples. The size of test sample is $10^4$.}
    \label{fig:pred_vs_label}
\end{figure}
Figure~\ref{fig:eff_vs_label} illustrates the testing accuracy as a function of input label for various numbers of excitatory neurons in a hidden layer. The left panel displays the overall accuracy, while the right panel shows the accuracy with ambiguity. The number of training samples per worker is $3 \times 10^4$. The color coding is consistent with Figure \ref{fig:eff_vs_nsample}.
Notably, the number of neurons in the hidden layer has a minimal impact on the overall test accuracy but significantly affects the test accuracy with ambiguity. For instance, the test accuracy with ambiguity for a network with 250 neurons trained directly is more than 10 times lower than that of a network with 10 neurons. With 250 neurons, the network trained through parallel training with diversity shows an accuracy with ambiguity approximately 2 times lower.
The highest overall accuracy is achieved for input label 0, which also has the lowest ambiguity. In contrast, input label 5 yields the lowest overall testing accuracy. The ambiguity increases from label 0 to label 5, then decreases at label 6, and subsequently increases again until label 9.

Figure \ref{fig:pred_vs_label} displays the distribution of incorrect predictions for various input labels using a 1-hidden layer network with 250 neurons (10 neurons per worker) trained through parallel training with diversity. Each worker received $3 \times 10^4$ training stimuli, and the test set consisted of $10^4$ samples. Several notable patterns emerge. For instance, the network most frequently confuses labels 4 and 9, as well as labels 3 and 5. Additionally, label 1 is often mispredicted as label 3 or 8,  Label 6 is frequently misclassified as label 0 and 5, while label 7 is often mispredicted as label 9 and 2.
\section*{Conclusions}
This study introduces a supervised learning approach for spiking neural networks that leverages spike-timing-dependent plasticity within a supervised framework for image recognition tasks, thereby eliminating the reliance on traditional backpropagation. The MNIST dataset effectively demonstrates the strengths of this approach. Using only 10 hidden neurons, the model obtains an 89\% accuracy rate with approximately 10\% ambiguity. When the model includes a larger number of hidden neurons and utilizes a diverse set of hyperparameters among them, it attains around 40\% learning accuracy with just 10 training stimuli, where each category is presented only once during training (one-shot learning). As the training sample size increases, the accuracy improves to 87\%, maintaining negligible ambiguity. Additionally,  a parallel training and testing method is proposed, leveraging small networks per computing process, which significantly reduces the training time and increases the testing accuracy with small training samples.  Future refinements, such as increasing the number of neurons in the first hidden layer,  introducing excitatory lateral connections,  or confining the region in feedforward synaptic connections, may further improve the network performance.

\section*{Acknowledgments}
The author would like to express sincere gratitude to Mark Linvill, Chris Orr, and the Department of Physics and Astronomy for facilitating access to HPC resources at the Purdue Rosen Center for Advanced Computing (RCAC). Thanks are also extended to the research service team at RCAC for their timely support. Moreover, the author would like to thank Marcel Stimberg and the Brian2 team for their invaluable discussions and assistance.


%
%
%
\bibliography{Ref}
\end{document}